\documentclass[10pt,twocolumn,letterpaper]{article}

\usepackage{cvm}
\usepackage{times}
\usepackage{epsfig}
\usepackage{graphicx}
\usepackage{amsmath}
\usepackage{amssymb}

\usepackage{booktabs}
\usepackage{multirow}
\usepackage[table]{xcolor}

\usepackage{cuted}
\usepackage{capt-of}

\def\etal{\emph{et~al.}}

\def\forexample{\emph{e.g.}}

\def\secmk{Sec.~}
\def\figmk{Fig.~}
\def\tablemk{Tab.~}
\def\equationmk{Eq.~}

\def\hmdo{HMDO~}

\def\onedot{.}
\def\etal{\emph{~et al}\onedot}

\usepackage[pagebackref=true,breaklinks=true,letterpaper=true,colorlinks,bookmarks=false]{hyperref}

\newcommand{\keywords}[1]{{\bf \emph{Keywords: #1}}}

\cvmfinalcopy 

\def\cvmPaperID{89} 

\ifcvmfinal\fi
\begin{document}


\title{\hmdo: Markerless Multi-view Hand Manipulation Capture with Deformable Objects}

\author{Wei Xie\footnotemark[1]\hspace{2mm}\hspace{5mm}
Zhipeng Yu\footnotemark[1]\hspace{2mm}\hspace{5mm}
Zimeng Zhao\hspace{2mm}\hspace{5mm} 
Binghui Zuo\hspace{2mm}\hspace{5mm} 
Yangang Wang\footnotemark[2]\\%
\\
Southeast University, China
}


\twocolumn[{
\renewcommand\twocolumn[1][]{#1}
\maketitle
\vspace{-8mm}
\begin{center}
   \includegraphics[width=1.0\textwidth]{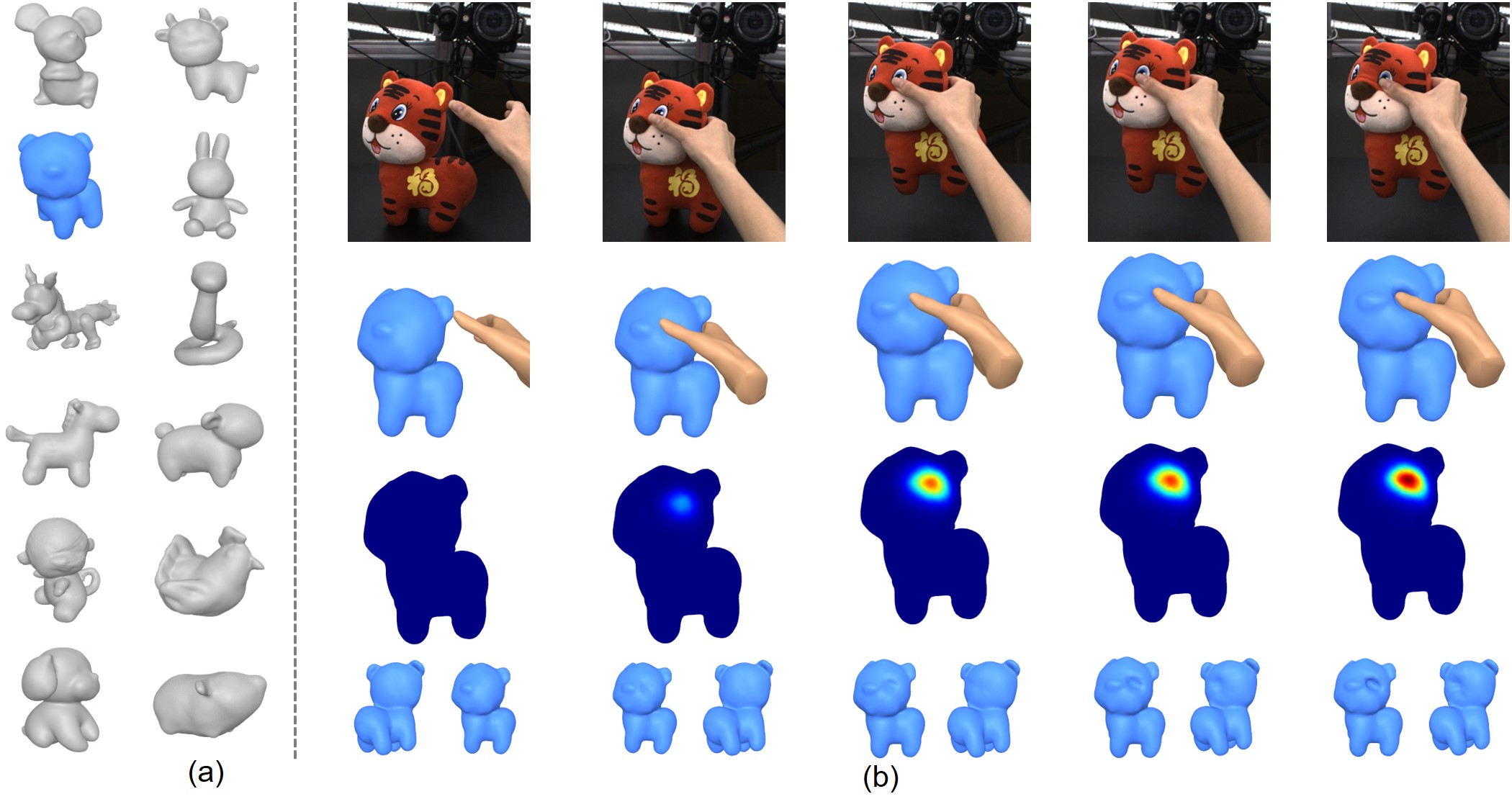}
   \captionof{figure}{\textbf{HMDO dataset}. (a) 12 zodiacs as manipulated objects; (b) Data including multi-view synchronized images, 3D meshes and contact deformation maps.}
   \label{fig0_teaser}
   \vspace{4mm}
\end{center}
}]

\begin{abstract}
   We construct the first markerless deformable interaction dataset recording interactive motions of the hands and deformable objects, called HMDO (Hand Manipulation with Deformable Objects). With our built multi-view capture system, it captures the deformable interactions with multiple perspectives, various object shapes, and diverse interactive forms. Our motivation is the current lack of hand and deformable object interaction datasets, as 3D hand and deformable object reconstruction is challenging. Mainly due to mutual occlusion, the interaction area is difficult to observe, the visual features between the hand and the object are entangled, and the reconstruction of the interaction area deformation is difficult.
   To tackle this challenge, we propose a method to annotate our captured data. Our key idea is to collaborate with estimated hand features to guide the object global pose estimation, and then optimize the deformation process of the object by analyzing the relationship between the hand and the object. Through comprehensive evaluation, the proposed method can reconstruct interactive motions of hands and deformable objects with high quality. HMDO currently consists of 21600 frames over 12 sequences.
   In the future, this dataset could boost the research of learning-based reconstruction of deformable interaction scenes. 
   \end{abstract}
   
   \keywords{Deformable interaction, Markerless capture, Mulit-view dataset, Collaborative reconstruction.}

\renewcommand{\thefootnote}{\fnsymbol{footnote}}
\footnotetext[1]{Contribute equally.} 
\footnotetext[2]{Corresponding author. E-mail: yangangwang@seu.edu.cn.}

\section{Introduction}
Understanding the interactive motions of hands and objects is an important topic in computer vision and graphics due to its wide range of applications in virtual reality, augmented reality, robotics, etc. In recent years, thanks to the development of deep learning and the creation of several hand and rigid object interaction datasets~\cite{hampali2020honnotate, chao2021dexycb, zhao2022stability, brahmbhatt2020contactpose,hampali2021handsformer,kwon2021h2o}, the research on the reconstruction of hands and rigid objects from monocular images has developed rapidly. However, these recent methods fail in the reconstruction of hands and deformable objects interactions. Mainly due to the need to solve non-rigid deformations of closely interacting contact regions and the lack of datasets for hand and deformable object interactions. Therefore, it is necessary to break the limitation of the lack of datasets that record hands and deformable objects.

To facilitate the study of data-driven methods to address scenarios where hands and deformable objects closely interact, we construct the first multi-view deformable interaction dataset. We built a multi-view capture system, using 10 high-speed industrial cameras to synchronously capture the close interaction process between hands and deformable objects at high frame rates from different perspectives. However, creating annotations for interactive motions of hands and deformable objects is very challenging. Reconstructing hand and deformable object interactions is more complicated than reconstructing hands and rigid objects interactions, because we not only need to solve the rigid motions of deformable objects, but also recover the non-rigid motions. In addition, due to occlusion, the interaction area is difficult to observe, and the visual features between the hand and the object are entangled, which makes it very difficult to reconstruct the deformation of the contact regions.

Most existing solutions~\cite{newcombe2015dynamicfusion, slavcheva2017killingfusion, guo2017real, zollhofer2014real,lin2022occlusionfusion} adopt fusion-based methods to reconstruct deformable objects. Some works propose specific strategies for interaction problems~\cite{zhang2019interactionfusion, zhang2021single}, they rely on extra depth cameras to record slow motions. Besides the depth dependence, these fusion-based methods~\cite{zhang2019interactionfusion, zhang2021single} to reconstruct interactions do not model the deformable objects explicitly, and could hardly obtain instance mesh sequences with time-invariant topology. Commonly, they could only tackle the interaction between single pair of hand-object. Other existing template-based methods~\cite{tsoli2018joint, petit2018capturing, frank2010learning, sengupta2020simultaneous, wuhrer2015finite, weiss2020correspondence} have difficulty in handling scenes where hands and objects are closely interacting. In our captured data, the human hands interact more closely with the deformable objects. This means that the interaction area is difficult to observe, the hand state needs to be reconstructed simultaneously, and the visual features between the hand and the object are entangled.

Aiming at the closely interactive motion reconstruction of hands and deformable objects, we propose a template-based method to annotate the data we captured. The method jointly reconstructs hand pose, object pose, and object deformation from captured data. All manipulated dolls are scanned and repaired in advance for their digital meshes with impermeable and homeomorphic properties. Our hand surface model also has higher resolution than MANO~\cite{romero2017embodied} and can more accurately represent the contact regions. The topological consistency of the object and hand meshes facilitates surface deformation analysis in our future studies. In terms of reconstruction algorithm design, the information of each viewing perspective is fully utilized to reconstruct the accurate hand, and the collaboration between the hand and the object is considered to guide the object pose estimation and the object deformation. Furthermore, a top-down strategy is adopted in our framework, so it theoretically supports reconstructing the interaction between multiple hands and deformable objects. Through comprehensive evaluation, the proposed method can reconstruct interactive motions of hands and deformable objects with high quality. 

The main contributions of this work are summarized as follows.

\noindent$\bullet$ A markerless deformable interaction dataset recording interactive motions between hands and deformable dolls of various appearances and morphology. The dataset will be publicly available on our website;

\noindent$\bullet$ A pipeline to reconstruct interactive motions of hands and deformable objects from multi-view data;

\noindent$\bullet$ An object deformation optimization algorithm under the guidance of hand and object collaboration.

\section{Related Work}
\subsection{Hand and Object Interaction Datasets}
In recent years, several datasets for hand and object interaction have already been proposed. ~\cite{garcia2018first} provided a dataset containing hand and object interactions, called FPHAB. They used a capture system consisting of magnetic sensors attached to the subject's hands and objects to obtain 3D annotations of the hands in RGB-D video sequences. However, since the magnetic sensors and the tape connecting them are visible, this changes the appearance of the hand in the color image. Hasson~\etal~\cite{hasson2019learning} introduced ObMan, which provided a large dataset of synthetic images of hands grasped objects. HO-3D~\cite{hampali2020honnotate} used several RGB-D cameras to capture sequences and presented the first markerless dataset of color images with 3D annotations for both the hand and object. Following this, DexYCB~\cite{chao2021dexycb} created a large-scale dataset of hands and rigid objects. Zhao~\etal~\cite{zhao2022stability} proposed a hand-object interaction dataset containing physical properties and stability metrics. ContactPose~\cite{brahmbhatt2020contactpose} proposed to use a thermal camera to capture the contact map of objects to reflect the common contact regions of the grasped objects. Hampali~\etal~\cite{kwon2021h2o} created a dataset containing 3D annotations of objects manipulated by two hands. However, existing hand and object interaction datasets only include interactions between hands and rigid objects, and lack data recording the interactions of hands and deformable objects. We present HMDO, the first markerless dataset recording interactive motions between hands and deformable dolls of various appearances and morphologies.

\begin{figure}[!t]
    \centering
    \includegraphics[width=\linewidth]{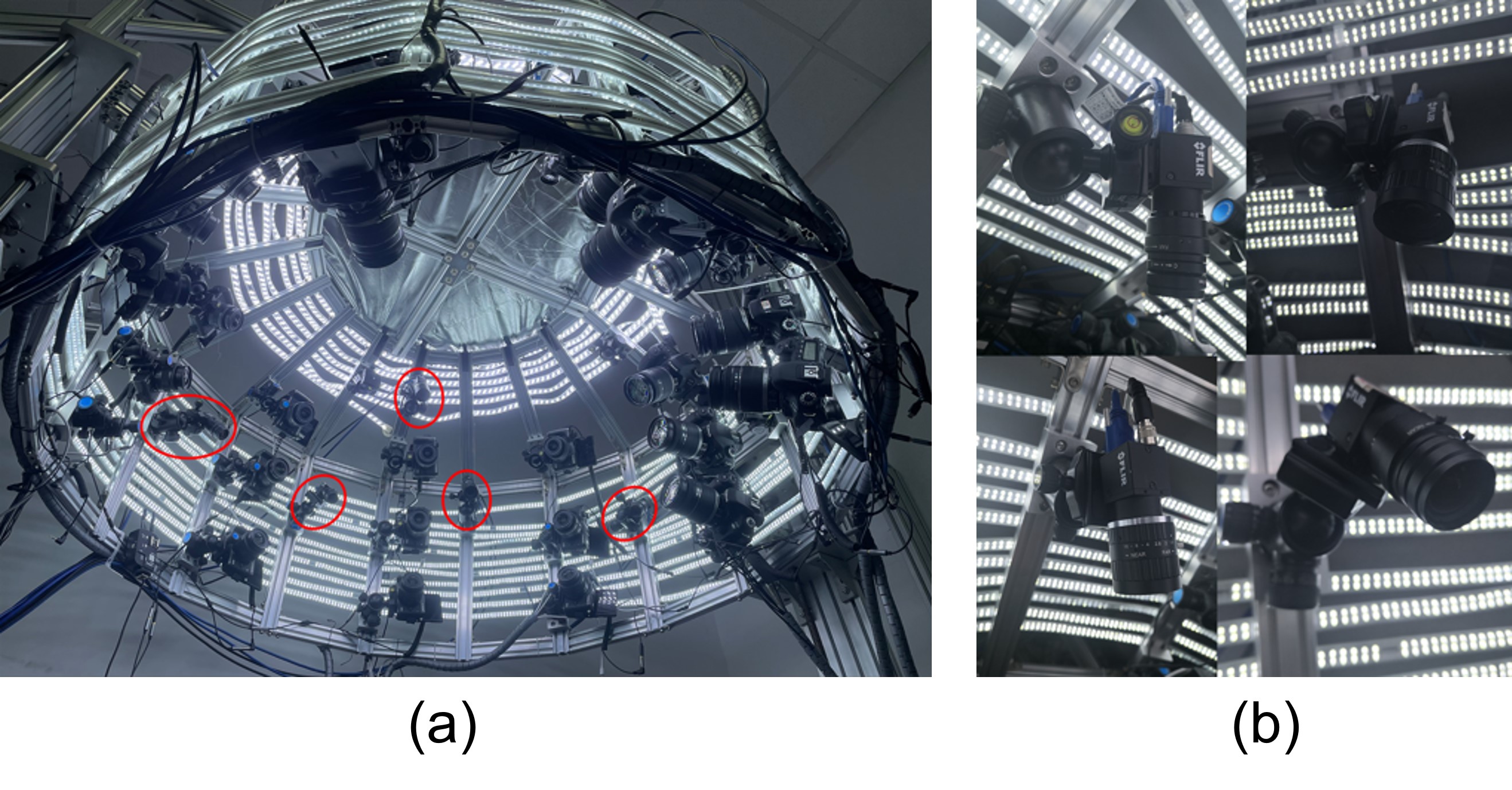}
    \caption{\textbf{Hardware system. } (a) Multi-view synchronized capture system; (b) Industrial cameras.}
    \vspace{-2mm}  
    \label{fig03_syn_capture_system}
\end{figure}

\subsection{Hand and Object Joint Reconstruction}
Existing work mainly focuses on hand and rigid object reconstruction. 
Because rigid objects only have global degrees of freedom (DoFs), the interaction between hand and rigid objects is easier to model, capture, and reconstruct. With the popularity of learning-based methods in the fields of graphics and 3D vision, datasets recording hand-only~\cite{zimmermann2019freihand, gomez2019large, zhao2020hand}, rigid-object-only~\cite{xiang2017posecnn, hinterstoisser2012model, tzionas20153d, liu2021stereobj}, and interaction between the two~\cite{garcia2018first,hasson2019learning, brahmbhatt2020contactpose, hampali2020honnotate,chao2021dexycb, zhao2022stability, kwon2021h2o} have increased rapidly in both quantity and quality. This further promotes more literature to adopt data-driven methods to solve related problems. Hasson~\etal~\cite{hasson2019learning} reconstructed the shape and pose of the hand-object through a unified network with extra synthetic data. \cite{hasson2020leveraging} proposed a sparsely supervised learning method to reconstruct hand-object, exploiting the photometric consistency between sparsely supervised frames. Cao~\etal~\cite{cao2021reconstructing} explored reconstructing hand-object interactions in the wild. Grady~\etal~\cite{grady2021contactopt} refine the estimated hand-object sate through contact prior learned from those datasets. Zhao~\etal~\cite{zhao2022stability} makes the interaction state more stable by introducing the optimization process based on a physics engine. The datasets recording hand interactions with rigid objects also speed up the comparison and evolution of methods for other problems including grasping generation~\cite{karunratanakul2020grasping, jiang2021hand} and manipulating planning~\cite{christen2021d, zhang2021manipnet, she2022learning}.

\subsection{Deformable Object Reconstruction}
Precisely, rigid objects are only ideal approximations, while deformable objects are more common in daily life, \forexample backpacks, clothes, dolls, and even human bodies. Reconstructing such objects has always been a difficult problem in the field. As one of the mainstream solutions, template-based methods often build deformable objects as finite element models (FEM), which are determined by object-specific parameters and have high DoF in the calculation. Some researchers~\cite{wuhrer2015finite,weiss2020correspondence} used sparse depth information to reconstruct deformation of single models. Some~\cite{tsoli2018joint, petit2018capturing} tried to generalize them to the scene containing weak hand-object interactions. Others~\cite{frank2010learning,sengupta2020simultaneous} obtain more deformation details by installing a depth camera and an extra force sensor on the robotic gripper at the same time. All these methods are difficult to generalize and apply to scenes where deformable objects interact closely with human hands. On the other hand, fusion-based methods~\cite{newcombe2015dynamicfusion, slavcheva2017killingfusion, innmann2016volumedeform, guo2017real} abandon explicit modeling of objects and reconstruct the entire scene as a field with slow changes in depth and illumination. Some hybrid attempts~\cite{zollhofer2014real,yu2015direct} divided the reconstruction process into two steps: online mesh template acquisition and real-time non-rigid reconstruction. Most of them take the human body, face, and hand as reconstruction objects, and do not consider the influence of occlusion. In recent years, some progresses~\cite{bozic2020deepdeform,bozic2020neural,lin2022occlusionfusion} have been made when combining with learning-based technologies to find correspondences between two adjacent camera frames. And the studies~\cite{zhang2019interactionfusion, zhang2021single} specific to interaction problems to distinguish hands and objects from the scene. Nevertheless, most of the above methods still aim at real-time performance rather than high-quality, high-precision, and topology-consistent dataset preparation. Therefore, to our best knowledge, there exist no large-scale datasets recording the interaction between human hands and deformable objects. When the hand interacts with the deformable object, the deformation of the hand is much smaller than that of the object. Therefore, most methods use the rigid approximation of the hand to take the deformation of the object as the main contradiction. We also adopt this assumption.

\begin{figure*}[!t]
    \centering
    \includegraphics[width=\linewidth]{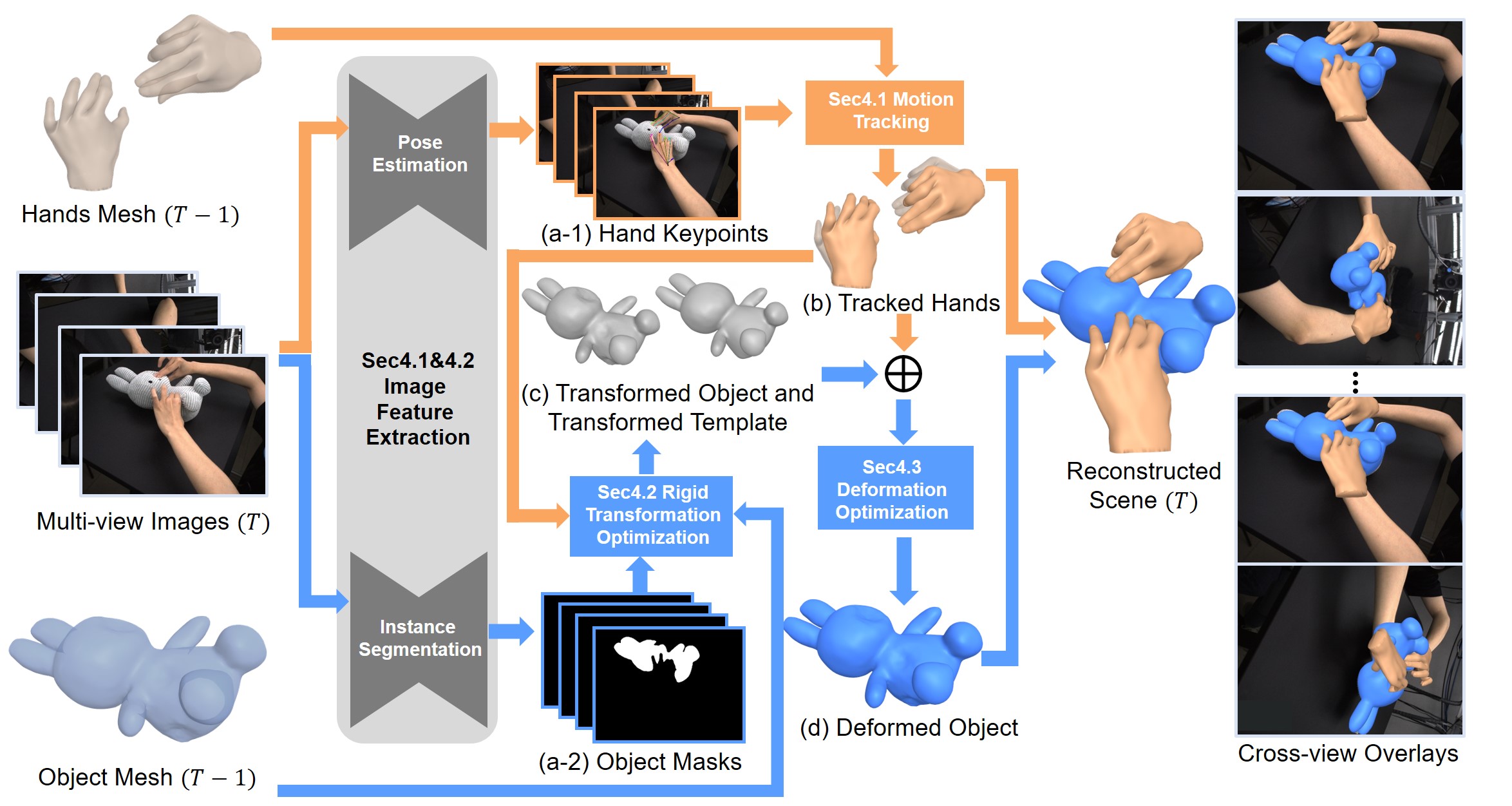}
    \caption{\textbf{Hand and deformable object reconstruction pipeline. }(a) Image features are first extracted from the current frame $T$, including the 2D poses of the hands and the masks of the object; (b) Hands motion tracking is performed based on image features and historical information; (c) The object global pose is estimated based on the current frame object masks and tracked hands, and the object mesh of the previous frame $T-1$; (d) The deformed object is obtained under the guidance of hand and object collaboration.}
    \vspace{-2mm}
    \label{fig02_pipeline}
\end{figure*}

\section{Data Capture}
\subsection{System Configuration}
We build a multi-view synchronized capture system to capture interactive motions of hands and deformable objects. The multi-view synchronized capture system uses hardware signals to trigger cameras shown in~\figmk\ref{fig03_syn_capture_system}. The system includes three components: an industrial camera array, a synchronous signal generator, and data caching devices.  Our camera array contains $10$ high-speed industrial cameras. Each one is equipped with $8mm$ focal lens and produces $2048\times1536$ resolution images. The capturing frequency and framed amount can be adjusted through Bluetooth before capturing. When capturing each hand manipulation motion, the signal generator sends hardware trigger signals to each camera through the Ethernet patch cable. It controls each camera to shoot with a negligible delay. Once the capture is finished, the data from the camera is transferred to the caching devices. This solution is capable of capturing stable data with a frame rate up to 110 FPS.

\subsection{Object Template Acquisition}
As shown in~\figmk\ref{fig0_teaser}, twelve zodiac dolls with various geometric and visual attributes are selected as the manipulation objects in our work. Although the doll comes from different manufacturers, we require that the elastic coefficient and density of the stuffing inside the doll should be as close as possible. In addition, during the data capture of hand-object interactions, we avoid pressure or tension that may cause plastic deformation for the dolls. We follow the steps to create a template object model for each doll. First, we adopt the fusion-based method~\cite{RecFusion} with a single RealSense435i device to capture a coarse object mesh, which shown as~\figmk\ref{fig11_template_acquisition}. Then a series of repairing processes including hole filling, isolated elements removal, and isotropic remeshing is adopted to guarantee that the whole object mesh is watertight and has genus 0 (homeomorphic to a sphere). After manual repair, these meshes with good geometric properties are used as template object models.

\section{Hand-Deformable Object Reconstruction}
An overview of our pipeline for reconstructing hand and deformable object interactions from multi-view motion data is shown in~\figmk\ref{fig02_pipeline}. Firstly, the method for hand tracking is described in~\secmk\ref{sec32_hand_motion_tracking}. Then, the method for object global pose estimation is given in~\secmk\ref{sec33_object_pose}. 
In~\secmk\ref{sec34_object_deformation_optimization}, we describe the optimization strategy for obtaining the deformable object by iteratively analyzing the relationship between hand and object. 
To facilitate the identification, for the object variables, the variables marked with hats superscript represent the results after global pose transformation, and the variables with tilde superscript represent the results after non-rigid deformation optimization. For the hand variables, the hand after optimization is represented by a tilde superscript.

\begin{figure*}[!t]
    \centering
    \includegraphics[width=\linewidth]{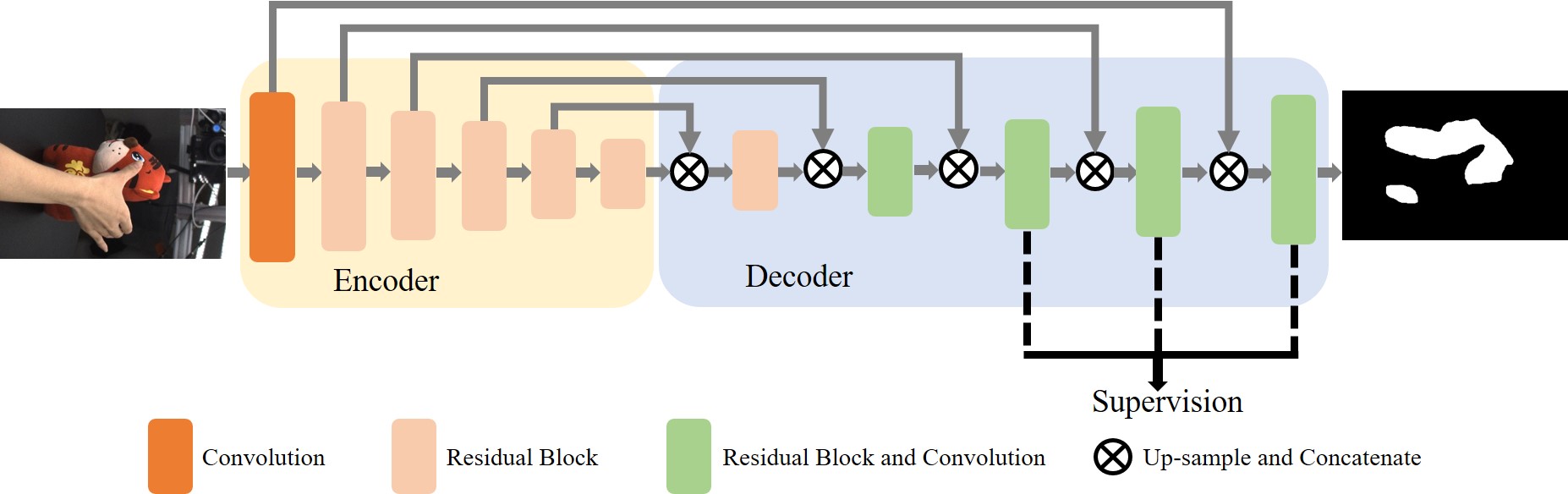}
    \caption{\textbf{Overview of object segmentation network. }We designed an object segmentation network, and the network used different scales of masks as supervision.}
    \vspace{-2mm}  
    \label{fig_object_segment_network}
\end{figure*}

\begin{figure}[!t]
    \centering
    \includegraphics[width=\linewidth]{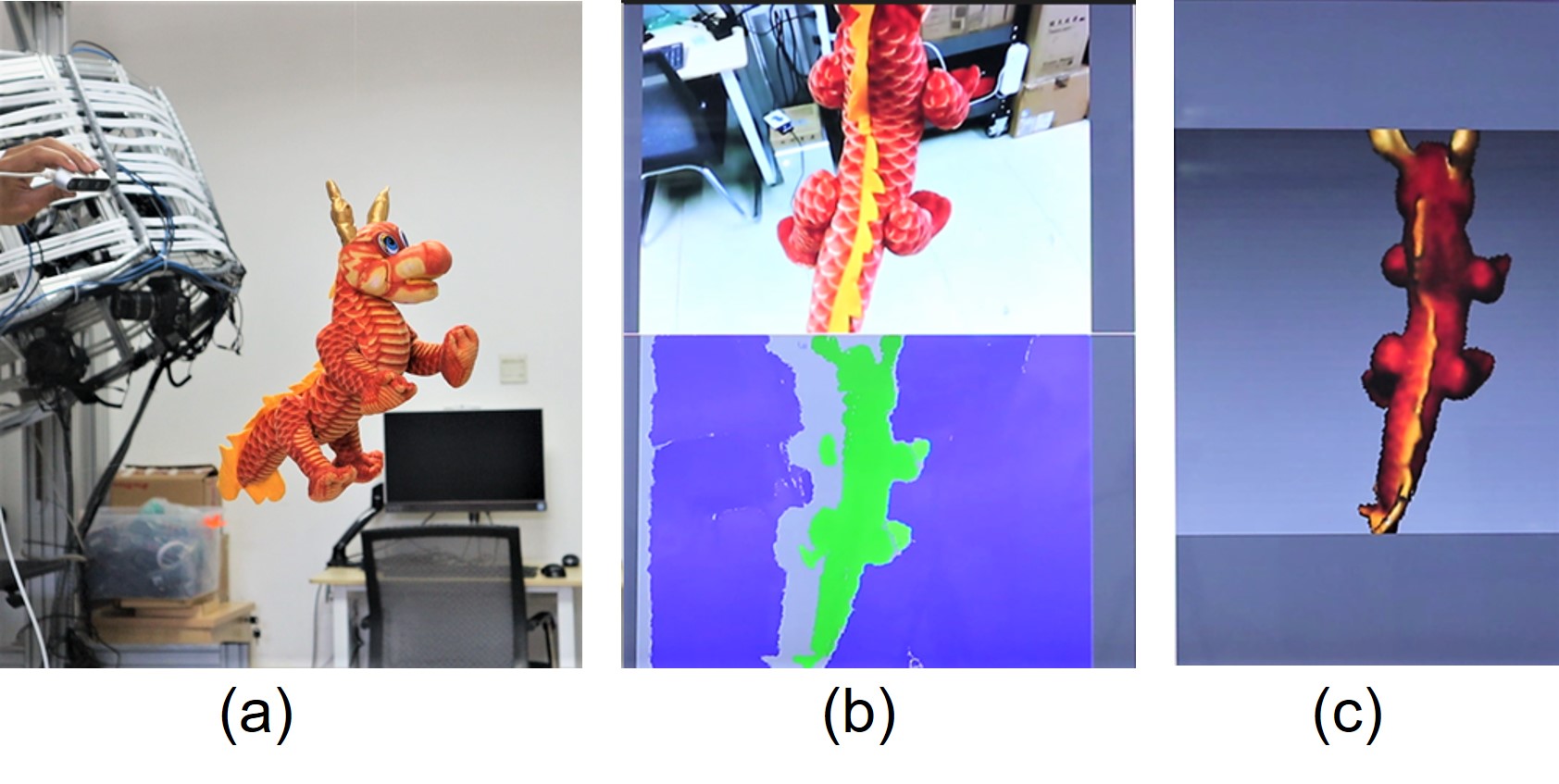}
    \caption{\textbf{Template acquisition. } (a) Reference object; (b) 
    Scanning process; (d) Reconstructed coarse template.}
    \vspace{-2mm}  
    \label{fig11_template_acquisition}
\end{figure}

\subsection{Hand Motion Tracking}
\label{sec32_hand_motion_tracking}

\noindent\textbf{Hand pose estimation. } First, we use the 2D pose estimation network~\cite{wang2019srhandnet} to estimate the 2D keypoints and confidences for each camera of the current frame. We then solve the following equation to obtain the 3D keypoints by utilizing camera parameters and the estimated 2D keypoints,
\begin{equation}
    \begin{aligned}
        \underset{k}{\operatorname{argmax}} \sum_{n \in \mathcal{N}}\left\|\left[X_{n}\right]_{\times} \cdot \Pi_{n}\left(R_{n} k+T_{n}\right)\right\|,
    \end{aligned}
    \label{eqn_hand3d}
\end{equation} 
where $\left[X_{n}\right]_{\times}$ is the skew matrix of the homogenous coordinate of $X_{n}$, which describes the 2D coordinate of hand keypoint in the $n$-th camera. $k$ is the 3D hand keypoint. $\Pi_{n}$ and $\left[R_{n}, T_{n}\right]$ are the intrinsic parameter and extrinsic parameter of the $n$-th camera, respectively. $\mathcal{N}$ is the set of valid camera views for this keypoint. It is worth noting that not all views have accurate 2D keypoints. For each camera, only when the confidence of the 2D keypoint is greater than a threshold, it will be regarded as a valid camera in the corresponding 3D keypoint calculation. In our experiment, the threshold was set to $0.6$.

\noindent\textbf{Hand mesh optimization.} We utilize the multi-view system to pre-optimize our hand surface model to obtain personalized hand shape parameters of subjects. During hand and deformable object tracking, only the pose $\boldsymbol{\theta}$ of the hand is optimized. We minimize the errors of deformed skeleton joints and estimated keypoints in both 3D and 2D space, 
\begin{equation}
    \begin{aligned}
        \underset{\boldsymbol{\theta}}{\operatorname{argmin}}\sum_{j}\left( \sum_{n \in \mathcal{N}}\left\|x_{j,n} - \mathcal{P}_{n}\left(f_{j}(\boldsymbol{\theta})\right)\right\| + 
        \left\|k_j - f_{j}(\boldsymbol{\theta})\right\| \right),
    \end{aligned}
    \label{eqn_hand_deform_skeleton}
\end{equation}
Where $x_{j,n}$ is the $j$-th 2D keypoint in the $n$-th camera, and $k_j$ is the $j$-th 3D keypoint. $f_{j}(\boldsymbol{\theta}) \in \mathbb{R}^{3\times1}$ represents the 3D position of the $j$-th hand skeleton joint with the parameter $\boldsymbol{\theta}$, and $\mathcal{P}_{n}(\cdot)$ can be expressed as:
\begin{equation}
    \begin{aligned}
        \mathcal{P}_{n}\left(f_{j}(\boldsymbol{\theta})\right) = R_nf_{j}(\boldsymbol{\theta}) + T_n.
    \end{aligned}
    \label{eqn_Proj_function}
\end{equation}

After getting the hand pose $\boldsymbol{\theta}$ of the current frame, We use historical information to do a temporal smoothing filter to obtain the smoothed pose $\boldsymbol{\tilde{\theta}}$. We adopt the linear blend skinning to deform our hand surface model $M_{h}$. Specifically, for the $i$-th vertex $v_{h}^{i}$ in the hand mesh, the deformed new position $\tilde{v}_{h}^{i}$ is computed as: 
\begin{equation}
    \begin{aligned}
        \tilde{v}_{h}^{i}=\sum_{j} \omega\left(v_{h}^{i}, f_{j}(\boldsymbol{\tilde{\theta}})\right)\left[R_{j}\left(v_{h}^{i}-f_{j}(\boldsymbol{\tilde{\theta}})\right)+f_{j}(\boldsymbol{\tilde{\theta}})+t_{j}\right]
    \end{aligned}
    \label{eqn_hand_deform_mesh}
\end{equation}   
where $\left[R_{j}, t_{j}\right]$ is the rigid transformation of the $j$-th bone, which is only determined by the hand pose $\boldsymbol{\tilde{\theta}}$, the skinning weight $\omega\left(v_{h}^{i}, f_{j}(\boldsymbol{\tilde{\theta}})\right)$ is computed by heat-based method, which measures the influence of the $j$-th bone to the $i$-th vertex.

\subsection{Object Pose Estimation}
\label{sec33_object_pose}
\noindent\textbf{Object segmentation. }We design an encoder-decoder network for object segmentation as shown in~\figmk\ref{fig_object_segment_network}. We use different scales of masks as supervision. Through this network, the object makes $\{S_{n}\}$ of the current frame can be obtained. $S_{n}$ represents the mask of the $n$-th view.

\noindent\textbf{Object pose optimization. }We use the genetic mutation algorithm to iteratively solve the global position of the object. In each iteration, we first select samples that are inherited to the next generation, and then we use the uniform distribution to simulate the mutation process based on the selected samples to obtain new samples for the next generation. Among them, we use the roulette method to determine the samples that can be inherited to the next generation. The smaller loss value of the sample, the greater the probability of inheritance to the next generation. The loss function for each sample is calculated as follows:
\begin{equation}
    \begin{aligned}
        L=\sum_{n} \mathcal{D}(M(\boldsymbol{\alpha}),\tilde{M}_{h}, S_{n}) + \lambda_{o}\|\boldsymbol{\alpha}\|_2
    \end{aligned}
    \label{eqn_individual_loss_term}
\end{equation} 
Where the first term is the reprojection error of the object, and the second term is the regularization term. $\boldsymbol{\alpha} \in \mathbb{R}^{6\times1}$ is the features of one sample. The first three dimensions represent the rotation, where we use axis-angle representation to describe the global rotation. The last three dimensions represent the translation. The first term can be expressed as:
\begin{equation}
    \begin{aligned}
        \mathcal{D}(M(\boldsymbol{\alpha}),\tilde{M}_{h}, S_{n})= 1-\frac{\mathcal{H}_{n}\left(\tilde{M}_{h},M(\boldsymbol{\alpha})\right)\cap S_{n}}{\mathcal{H}_{n}\left(\tilde{M}_{h},M(\boldsymbol{\alpha})\right)\cup S_{n}}
    \end{aligned}
    \label{eqn_individual_loss_first_term}
\end{equation}
where, $\tilde{M}_{h}$ is the hand mesh of the current frame we obtained in~\ref{sec32_hand_motion_tracking}, and $M(\boldsymbol{\alpha})$ represents the object mesh with the parameter $\boldsymbol{\alpha}$, i.e., global rigid transformation. $\mathcal{H}_{n}\left(\tilde{M}_{h},M(\boldsymbol{\alpha})\right)$ is the rendered mask of the object mesh in the $n$-th camera that is not occluded by the hand mesh. We use the global pose of the previous frame as the initial one in our optimization.



Our goal is to minimize the overall loss, which is the sum of $L$ for all samples. Through iterative optimization, we obtain the object global pose that converges to the optimal solution. We then use the historical information to perform a temporal smoothing filter on the optimized pose. Since our method is a global optimization strategy, it does not accumulate temporal errors.

\begin{figure}[!t]
    \centering
    \includegraphics[width=\linewidth]{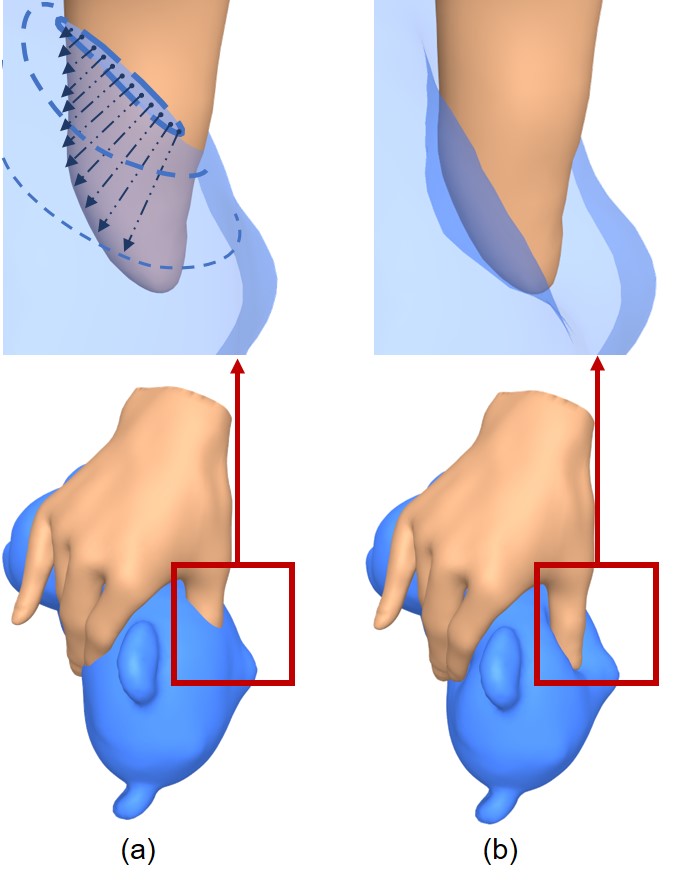}
    \caption{\textbf{Deformation schematic diagram. }(a) Result before  deformation;  (b) Result after deformation.}
    \vspace{-2mm}
    \label{fig10_deformation_diagram}
\end{figure}

\subsection{Object Deformation Optimization}
\label{sec34_object_deformation_optimization}
To model non-rigid deformation of deformable objects, we follow the representation in embedded deformation graphs~\cite{sumner2007embedded}. The non-rigid deformation is represented by the affine transformation $\{A_{s}, t_{s}\}$ of uniformly selected vertices $\{g_{s}\}$ on the mesh, which are treated as nodes of the graph. For the vertex $\hat{v}_{o}$ in the globally transformed object template mesh, the new position of the deformation $\tilde{v}_{o}$ is computed as:
\begin{equation}
    \begin{aligned}
        \tilde{v}_{o} =\sum_{g_{s} \in \mathcal{S}\left(\hat{v}_{o}\right)} \omega\left(\hat{v}_{o}, g_{s}\right)\left[A_{s}\left(\hat{v}_{o} -g_{s}\right)+g_{s}+t_{s}\right]
    \end{aligned}
    \label{eqn_object_deform_mesh}
\end{equation}   
where $\mathcal{S}(\hat{v}_{o})$ is the neighbor nodes of mesh vertex $\hat{v}_{o}$. $\omega(\hat{v}_{o}, g_{s})$ is the deformation weights of node $g_{s}$ to $\hat{v}_{o}$, which represents the influence of the node on the vertex. The definition of vertex neighbor nodes and the calculation of deformation weights refer to~\cite{sumner2007embedded}.
To get the transformation $\{A_{s}, t_{s}\}$ of all nodes in the deformation graph, we estimate them by optimizing the following function:
\begin{equation}
    \begin{aligned}
        E=\lambda_{1}E_{\text {cont}}+\lambda_{2}E_{\text {silh}}+\lambda_{3}E_{\text {temp}}+\lambda_{4}E_{\text {rigid}}+\lambda_{5} E_{\text {reg}}.
    \end{aligned}
    \label{eqn_object_energy_function}
\end{equation}

\noindent\textbf{Contact term. }$E_\text {cont}$ is the contact term. By analyzing the relationship between the hand and the object, the object is deformed to match the contact, resulting in a reasonable manipulation. This term can be expressed as:
\begin{equation}
    \begin{aligned}
        E_{\text {cont}}=\sum_{i} ||\tilde{v}_{o}^{i}-{v}_{target}^{i}||.
    \end{aligned}
    \label{eqn_object_contact_term}
\end{equation}  
where $\tilde{v}_{o}^{i}$ is the deformed position of $\hat{v}_{o}^{i}$, and ${v}_{target}^{i}$ is the target position of $\hat{v}_{o}^{i}$ by analyzing the relationship between the hand and the object. The object vertices target position are obtained by the following steps. 
After rigid transformation of the current frame object in~\secmk\ref{sec33_object_pose}, rays are emitted from the object to the hand for intersection detection. If penetration occurs, we record the vertex on the object, as well as the first intersection with the hand, which are marked as $\left(\hat{p}_{o}, \tilde{p}_{h} \right)$. The standard 3D Axis-Aligned Bounding Box~\cite{alliez20123d} tree is used to speed up the process.
We set the target position of these penetrated vertex $\{\hat{p}_{o}\}$ to $\{\tilde{p}_{h}\}$. The regions squeezed by the hand affect the surrounding regions. We employ a strategy based on geodesic distance and penetration depth to diffuse deformation around. The target positions of the remaining vertices can be expressed as:  
\begin{equation}
    \begin{aligned}
        {v}_{target} = \hat{v}_{o} - \frac{1}{N} \sum_{i=0}^N \mathcal{I}(\hat{v}_{o},{\hat{p}_{o}}^{i}) * \vec{n} 
    \end{aligned}
    \label{eqn_vpen}
\end{equation}
where, $\hat{v}_{o}$ is the vertex position of the object before deformation, and $N$ represents the number of penetrated vertex affecting vertex $\hat{v}_{o}$. $\vec{n}$ is the unit normal vector of vertex $\hat{v}_{o}$. $\mathcal{I}(\cdot)$ is the impact factor, which can be expressed as:
\begin{equation}
    \begin{aligned}  
        \mathcal{I}(\hat{v}_{o},\hat{p}_{o}) = d\left(\hat{p}_{o}\right) * \mathrm{exp}( -\lambda_{c} * \mathcal{G}(\hat{v}_{o}, \hat{p}_{o}))
    \end{aligned}
    \label{eqn_influence_factor}
\end{equation}
where, $d\left(\hat{p}_{o}\right)$ represents the penetration depth of vertex $\hat{p}_{o}$, and $\mathcal{G}(\cdot)$ is the geodesic distance. Regarding the calculation of geodesic distance, we refer to~\cite{crane2013geodesics}. In our experiments, when $\mathcal{I}(\cdot)$ is less than 0.02, we consider the vertex $\hat{v}_{o}$ are not affected by the vertex $\hat{p}_{o}$.

\noindent\textbf{Silhouette term. }The $E_{\text {silh}}$ term constrains the projection of the object model under each camera perspective to be consistent with the contour of the observed images.
\begin{equation}
    \begin{aligned}
        E_{\text {silh}}=  \sum_{\hat{v}_{o} \in \mathcal{C}} \sum_{n \in \mathcal{N}\left(\hat{v}_{o}\right)} \left\|\mathcal{P}_{n}\left(\tilde{v}_{o}\right) - c_{\hat{v}_{o},n}\right\|
    \end{aligned}
    \label{eqn_object_fit_term}
\end{equation}
where $\mathcal{C}$ includes all the vertices that have corresponding contours in the observed images. $\mathcal{N}\left(\hat{v}_{o}\right)$ records all camera numbers for which $\hat{v}_{o}$ has corresponding contour points in the observed camera perspective. $c_{\hat{v}_{o},n}$ is the corresponding contour point of vertex $\hat{v}_{o}$ under the observed image of $n$-th camera. $\tilde{v}_{o}$ is the deformation result of $\hat{v}_{o}$, and $\mathcal{P}_{n}\left(\tilde{v}_{o}\right)$ is the 2d projection of $\tilde{v}_{o}$. For methods of finding the matching 2d pixels on the image plane and retrieving the 3D position of 2D image coordinates, we refer to~\cite{wang2020personalized}.

\noindent\textbf{Temporal smoothing term.} The temporal smoothing term $E_{\text {temp}}$ encourages smooth deformation from frame to frame, which can be expressed as:
\begin{equation}
    \begin{aligned}
        E_{\text {temp}}=\sum_{i} ||\tilde{v}_{o}^{i}-\hat{v}_{last}^{i}||.
    \end{aligned}
    \label{eqn_object_temporal}
\end{equation}  
where $\tilde{v}_{o}^{i}$ is the deformed position of $\hat{v}_{o}^{i}$, and $\hat{v}_{last}^{i}$ is the corresponding vertex on the object mesh of the previous frame after the global transformation in~\secmk\ref{sec33_object_pose}.

\noindent\textbf{Rigid term. }The term $E_{\text {rigid}}$ used to restrict the affine transformation to be as rigid as possible, which is the same~\cite{sumner2007embedded} and is formulated as:
\begin{equation}
    \begin{aligned}
        E_{\text {rigid }} =&\sum_{s}\left(\left(\mathbf{a}_{s, 1}^{T} \mathbf{a}_{s, 2}\right)^{2}+\left(\mathbf{a}_{s, 1}^{T} \mathbf{a}_{s, 3}\right)^{2}+\left(\mathbf{a}_{s, 2}^{T} \mathbf{a}_{s, 3}\right)^{2}\right)  \\
    &+\sum_{s}\sum_{i}\left(\left(1-\mathbf{a}_{s, i}^{T} \mathbf{a}_{s, i}\right)^{2}\right),
    \end{aligned}
    \label{eqn_object_rigid_term}
\end{equation}
where $\mathbf{a}_{s, 1}$, $\mathbf{a}_{s, 2}$ and $\mathbf{a}_{s, 3}$ are the column vectors of $A_{s}$. 

\noindent\textbf{Regularization term. }The term $E_\text {reg}$ is served as a regularizer for the deformation by indicating that the affine transformations of adjacent graph nodes should agree with one another. Specifically, it means that the affine transformation of node $g_{m}$ is applied to node $g_{n}$, which should be consistent with the affine transformation of node $g_{n}$ being applied to itself.
\begin{equation}
    \begin{aligned}
        E_{\text {reg}}=&\sum_{m} \sum_{g_{n} \in \mathcal{S}\left(g_{m}\right)} \omega\left(g_{n}, g_{m}\right)|| g_{m} + t_{m} \\
        + &A_{m}\left(g_{n}-g_{m}\right) -\left(g_{n}+t_{n}\right)||.
    \end{aligned}
    \label{eqn_object_reg_term}
\end{equation}  

We optimize the function~\equationmk\ref{eqn_object_energy_function} and update the object mesh at the end of each iteration. The deformation schematic diagram is shown in~\figmk\ref{fig10_deformation_diagram}.

\begin{figure*}[!t]
    \centering
    \includegraphics[width=\linewidth]{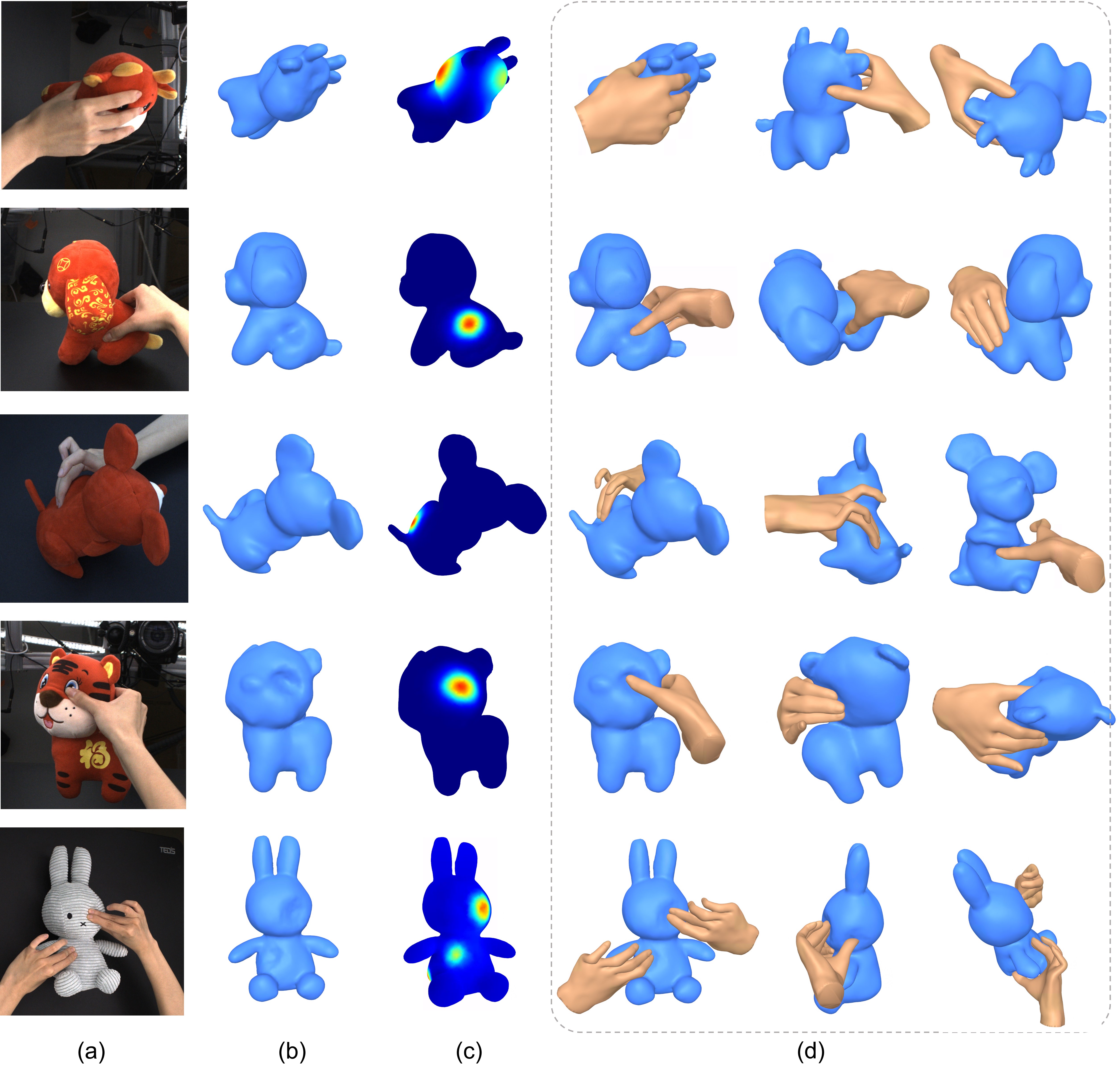}
    \vspace{-4mm}
    \caption{\textbf{Qualitative results of our method on different objects. }Randomly selected frames from hands and objects interaction sequences. (a) Input frames; (b) Object meshes; (c) Contact deformation maps; (d) Hand and object meshes (3 views).}
    \vspace{1mm}
    \label{fig09_results}
\end{figure*}

\section{Experiments}
\subsection{Implementation Details}
All of our experiments are performed on a computer configured with an Intel i7- 12700 CPU, NVIDIA GeForce RTX 3090 GPU. Our hand surface model has 6829 vertices and 13654 faces. The number of vertices of our object templates is between 6000 and 13000, and the number of faces is between 12000 and 25000. 

\noindent\textbf{Hand pose estimation. }The architecture of our 2D hand pose estimation network is based on SRNet~\cite{wang2019srhandnet}. We use the existing hand and rigid object interaction datasets HO-3D~\cite{hampali2020honnotate}, DexYCB~\cite{chao2021dexycb}, CBF~\cite{zhao2022stability} and ContactPose~\cite{brahmbhatt2020contactpose} as our training dataset. During network training, we use the SGD optimizer with the learning rate set to $10^{-5}$. After training, we collect data in our synchronized multi-view capture system, estimate 2D hand pose from these data using the trained model, and manually adjust for the incorrect poses. We then fine-tune the hand 2D pose estimation model on these adjusted data. The 3D pose is estimated from the multi-view 2D keypoints, and we use the LDLT method to solve it. The non-linear optimization of hand surface model is solved by Ceres~\cite{agarwal2015others}.

\noindent\textbf{Object mask segmentation. }During network training, we use the SGD optimizer, the network learning rate is set to $10^{-5}$, and the MSE loss is used for supervision. Regarding the training data, for each object, we first use a depth camera to collect coarse data through distance threshold segmentation, and we use color threshold segmentation and manual processing to obtain accurate masks. In addition, We perform data augmentation on these data, including random combinations with data from existing hand datasets, geometric transformations, and color transformations of the images.

\noindent\textbf{Object global pose optimization. }In the first frame, we use the uniform distribution to initialize the samples, the population size of samples is set to 500, and the number of iterations is set to 20. In subsequent frames, initial population distribution and population size remain the same, but only 1 iteration is performed.

\noindent\textbf{Object deformation. }The weight $\lambda_{c}=0.2$, $\lambda_{1}=5$, $\lambda_{2}=5$,$\lambda_{3}=1$, $\lambda_{4}=1$, $\lambda_{5}=2$. The non-linear optimization of object deformation is solved by Ceres~\cite{agarwal2015others}.

\subsection{Qualitative Results}
\begin{figure}[!t]
    \centering
    \includegraphics[width=\linewidth]{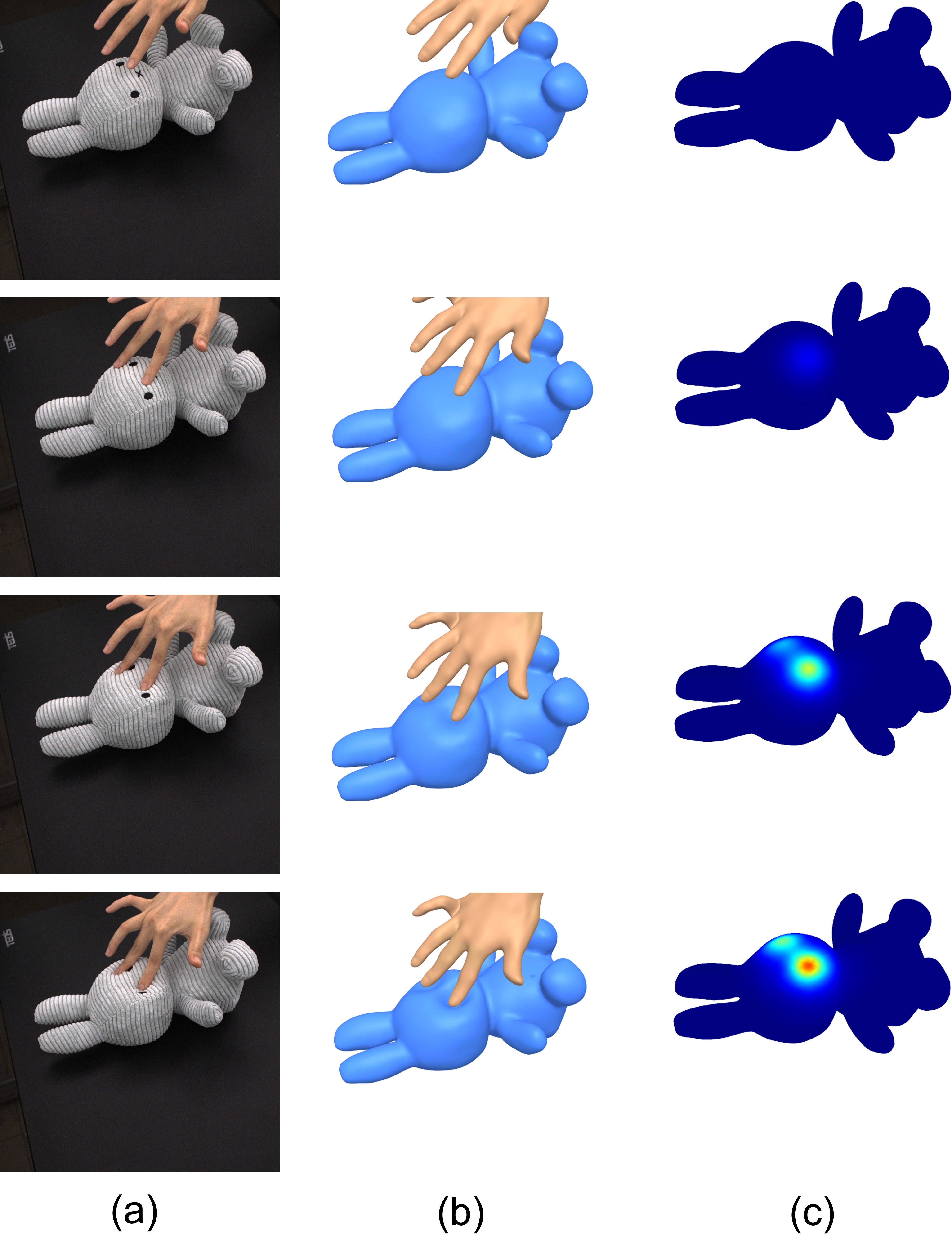}
    \caption{\textbf{Reconstruction results on the sequence "Rabbit". }(a) Selected frames; (b) Reconstruction results; (c) Contact deformation distribution maps.}
    \label{fig08_rabbit_sequence}
\end{figure}

\begin{figure}[!t]
    \centering
    \includegraphics[width=\linewidth]{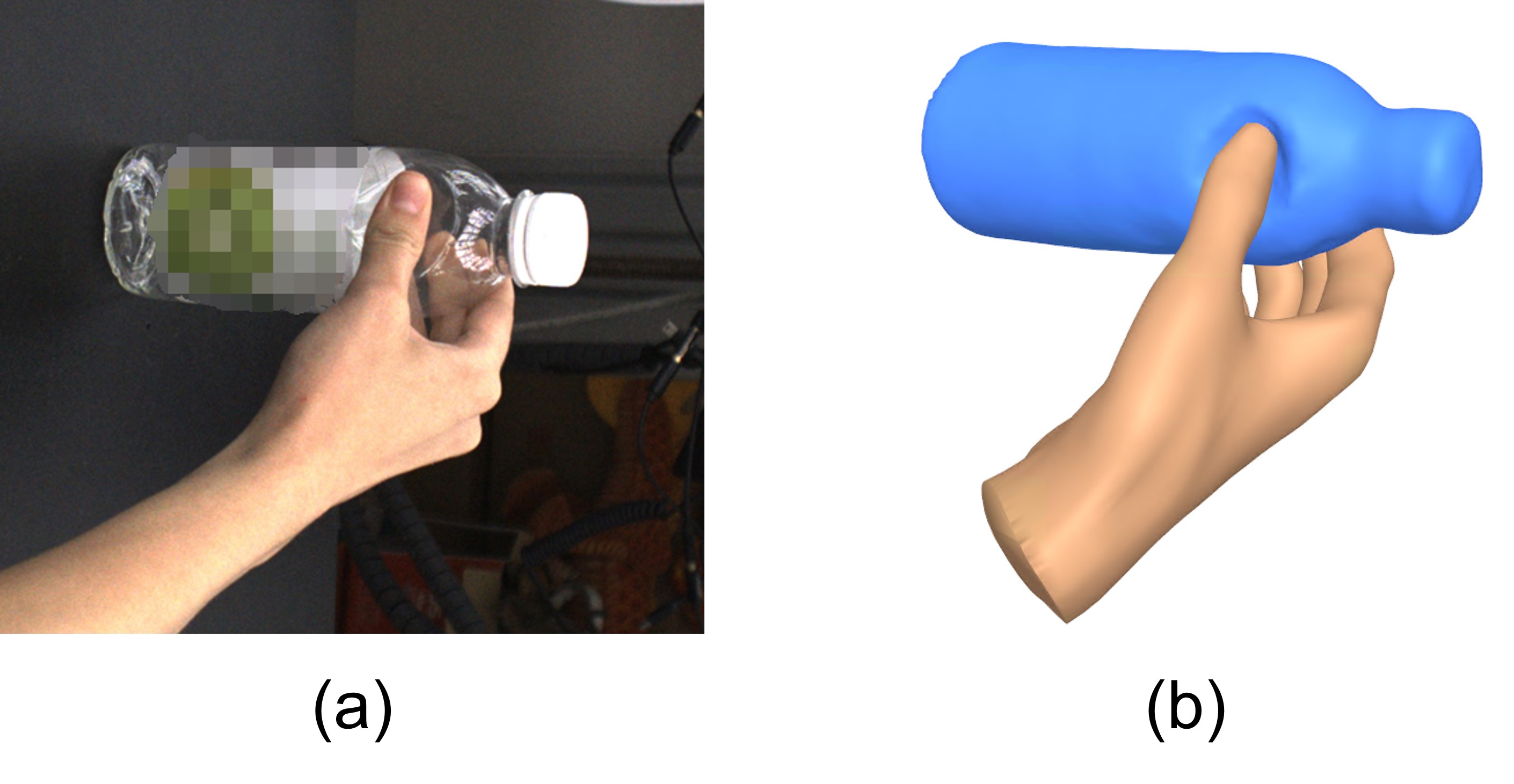}
    \caption{\textbf{Reconstruction result of plastic water bottle}. (a) Test frame; (b) Reconstruction result.}
    \label{fig12_non_doll_objects}
\end{figure}

We show the reconstruction results from the captured motion data in~\figmk\ref{fig0_teaser}, \figmk\ref{fig08_rabbit_sequence} and \figmk\ref{fig09_results}. In~\figmk\ref{fig0_teaser}, we show the interaction between the hand and the tiger. In~\figmk\ref{fig08_rabbit_sequence}, we show the interaction between the hand and the rabbit. We capture at a high frame rate and there is less variation from frame to frame, so to reflect the difference, the displayed frames are shown at specific time intervals. \figmk\ref{fig09_results} shows randomly selected frames in the interactive motion of the hand and different objects. From these reconstruction results, we can see that our solution is able to track and reconstruct the interaction of hands and deformable objects in various poses with high quality. Manipulated objects can vary greatly in visual and shape. 
It is worth mentioning that our method can handle multiple hands interacting with objects, as our framework adopts a top-down strategy. In~\figmk\ref{fig09_results} we show the interaction result between hands and rabbits.
In addition, in~\figmk\ref{fig12_non_doll_objects}, we show the reconstruction of the hand and plastic water bottle interactions using the proposed method.

\begin{table}[t]
    \begin{center}
        \resizebox{1\linewidth}{!}{
    \begin{tabular}{c|cccc}
    \noalign{\hrule height 1.5pt}
    \rowcolor{white}Methods 
    &4 views  &6 views &8 views &10 views\\  
     
    \midrule
    Mean.$(\mathrm{mm})$ 
        &14.75 &10.66 &7.31  &5.58 \\
    Std.$(\mathrm{mm})$
    &6.81 &4.53 &3.64  &2.29 \\

    \noalign{\hrule height 1.5pt}
    \end{tabular}
    }
    \end{center}

    \caption{\textbf{Evaluations the accuracy of hand pose estimation. } We report the average hand joint errors for different camera number settings. }
    \label{tab01_evaluation_hand}
\end{table}

\subsection{Evaluation of Hand 3D Pose Estimation}
Estimating accurate hand pose is important for hand motion tracking, while also correctly guiding rigid transformations and non-rigid deformations of objects. To quantitatively evaluate the accuracy of the hand pose estimated by our annotation method, we manually annotated the 3D locations of the 3D joints in randomly selected frames.~\tablemk\ref{tab01_evaluation_hand} shows the estimated hand pose accuracy for different numbers of views. From the results in~\tablemk\ref{tab01_evaluation_hand}, as the number of camera views increases, the estimation error gradually decreases. We can achieve an average joint error accuracy of lower than 6mm on average with all camera view settings.

\begin{table}
    \begin{center}
        \resizebox{1\linewidth}{!}{
          \begin{tabular}{c| ccc}
                \noalign{\hrule height 1.5pt}
                Parameters & mIoU.$(\mathrm{\%})\uparrow$ & Time. $(\mathrm{s})\downarrow$ \\
                \midrule
                size = 100, iter = 20, $\mathcal{U}$ & 83.33  & 17.72 \\
                size = 300, iter = 20, $\mathcal{U}$ & 84.72  & 49.76 \\
                size = 500, iter = 20, $\mathcal{U}$ & 86.91  & 82.70 \\
                size = 700, iter = 20, $\mathcal{U}$ & 86.90  & 123.59 \\
                \midrule
                size = 500, iter = 5, $\mathcal{U}$ & 58.36  & 20.72 \\
                size = 500, iter = 10, $\mathcal{U}$ & 82.98  & 43.27 \\
                size = 500, iter = 20, $\mathcal{U}$ & 86.91  & 82.70 \\
                size = 500, iter = 30, $\mathcal{U}$ & 86.79  & 120.18 \\
                \midrule
                size = 500, iter = 20, $\mathcal{O}$  & 84.74  & 84.47 \\
                size = 500, iter = 20, $\mathcal{U}$ & 86.91  & 82.70 \\
                \noalign{\hrule height 1.5pt}
          \end{tabular}
        }
    \end{center}
    \caption{\textbf{Evaluation of object pose estimation accuracy in initial frames with different parameter settings. }$\mathcal{U}$ represents uniform distribution, and $\mathcal{O}$ represents normal distribution. ``size'' represents population size, and ``iter'' represents optimization iterations number.}
    \label{tab02_effectobjectpose_initialframe}
\end{table}

\subsection{Evaluation of Object Pose Estimation}
The population size, initial population distribution, and optimization iterations number can affect the performance of our object global pose estimation algorithm. To evaluate the impact of these parameters on object pose estimation, we conduct qualitative and quantitative experiments. For experiments, we manually annotated object masks in 2 sequences. In these evaluation samples, the objects have no non-rigid deformations. We use time-consuming and the mean intersection over union(mIoU) to evaluate the performance of object pose estimation under different parameter settings. The mIoU is the mean of ratio of the intersection and union of the predicted value and the true value. We measure the difference between the 
rendered mask of the object mesh and ground-truth.

\begin{table*}
    \begin{center}
        \resizebox{1\linewidth}{!}{
          \begin{tabular}{c| cccccc}
                \noalign{\hrule height 1.5pt}
                Terms & Initialization &  $E_{\text {cont}}$ & $E_{\text {cont}}$ + $E_{\text {reg}}$ &  $E_{\text {cont}}$ + $E_{\text {reg}}$ + $E_{\text {rigid}}$ &  $E_{\text {cont}}$ + $E_{\text {reg}}$ + $E_{\text {rigid}}$ + $E_{\text {temp}}$ &  $E_{\text {cont}}$ + $E_{\text {reg}}$ + $E_{\text {rigid}}$ + $E_{\text {temp}}$ + $E_{\text {silh}}$ 
                \\
                \midrule
                mIoU.$(\mathrm{\%})\uparrow$ & 78.54 & 83.26 & 84.92 & 85.44 & 85.61 & 87.53 \\
                Inter.$(\mathrm{cm}^3)\downarrow$ & 15.10 & 6.13 & 4.05 & 3.52 & 3.64 & 3.27 \\
                \noalign{\hrule height 1.5pt}
          \end{tabular}
        }
    \end{center}

    \caption{\textbf{Evaluation of terms for object deformation optimization.} ``Initialization'' denotes the object mesh before deformation.}
    \label{tab04_ablation_object}
\end{table*}

\begin{figure}[!t]
    \centering
    \includegraphics[width=\linewidth]{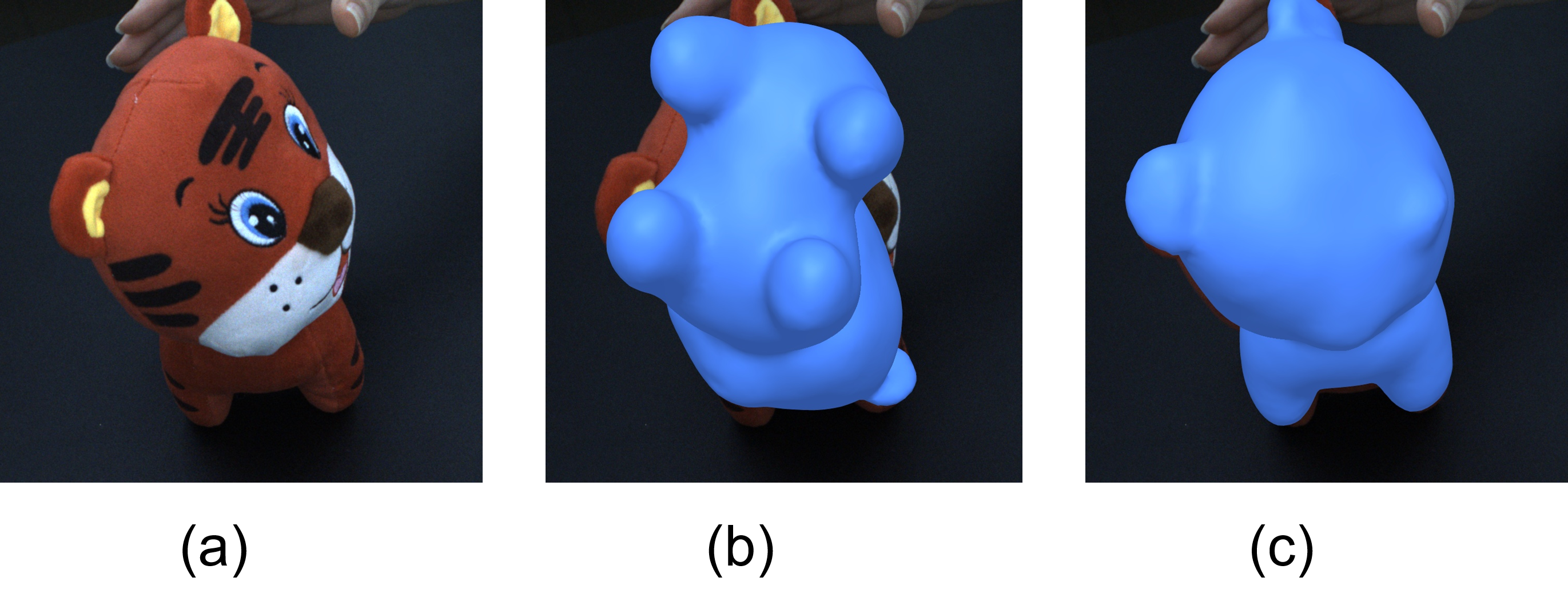}
    \caption{\textbf{Incorrect object pose estimation}. (a) Test frame; (b) Result with incorrect pose; (c) Result with correct pose. Inappropriate parameter settings are easy to fall into local optimal solution, resulting in incorrect object pose estimation.}
    \vspace{-2mm}
    \label{fig04_incorrect_pose}
\end{figure}

We first treat these data as independent initial frames and optimize them to obtain object pose. The effect of different population size is compared in the first 4 rows of \tablemk\ref{tab02_effectobjectpose_initialframe}. When the population size is set to 500, similar results can be obtained with the population size set to 700, and the time is shorter. The middle four rows of \tablemk\ref{tab02_effectobjectpose_initialframe} evaluate the effect of the number of iterations on the results. Satisfactory results can be achieved with 20 iterations. The experiments on the effect of the initial population distribution on the results are shown in the last two rows of \tablemk\ref{tab02_effectobjectpose_initialframe}. It can be seen from the table that the initial population set to uniform distribution has higher accuracy of pose estimation than a normal distribution. This is because the initial frame has no prior information. In other words, there is no guidance for the initial pose, so using uniform distribution is less likely to fall into local optimal solutions than a normal distribution. As shown in~\figmk\ref{fig04_incorrect_pose}, inappropriate parameter settings are easy to fall into the local optimal solution, resulting in incorrect object pose estimation.

\begin{table}
    \begin{center}  
        \resizebox{1\linewidth}{!}{
          \begin{tabular}{c| ccc}
                \noalign{\hrule height 1.5pt}
                Parameters & mIoU.$(\mathrm{\%})\uparrow$ & Time. $(\mathrm{s})\downarrow$ \\
                \midrule
                size = 500, iter = 1, $\mathcal{U}$  & 84.63  & 4.17 \\
                size = 500, iter = 3, $\mathcal{U}$  & 84.82  & 12.53 \\
                \noalign{\hrule height 1.5pt}
          \end{tabular}
        }
    \end{center}
    \caption{\textbf{Evaluation of object pose estimation accuracy in non-initial frames with different parameter settings. }$\mathcal{U}$ represents uniform distribution. ``size'' represents population size, and ``iter'' represents optimization iterations number.}
    \label{tab03_effectobjectpose_non-initialframe}
\end{table}

For non-initial frames, we use the result of the previous frame as the initial value to optimize the object pose. As we collect data at a high frame rate, there is less variation from frame to frame. We evaluate the effects of iteration number and initial population distribution on non-initial frame pose estimation with two sequences that are manually annotated. As shown in \tablemk\ref{tab03_effectobjectpose_non-initialframe}, setting the number of iterations to 1 can quickly converge to a suitable solution.

\subsection{Evaluation of Object Deformation Optimization}
\begin{figure}[!t]
    \centering
    \includegraphics[width=\linewidth]{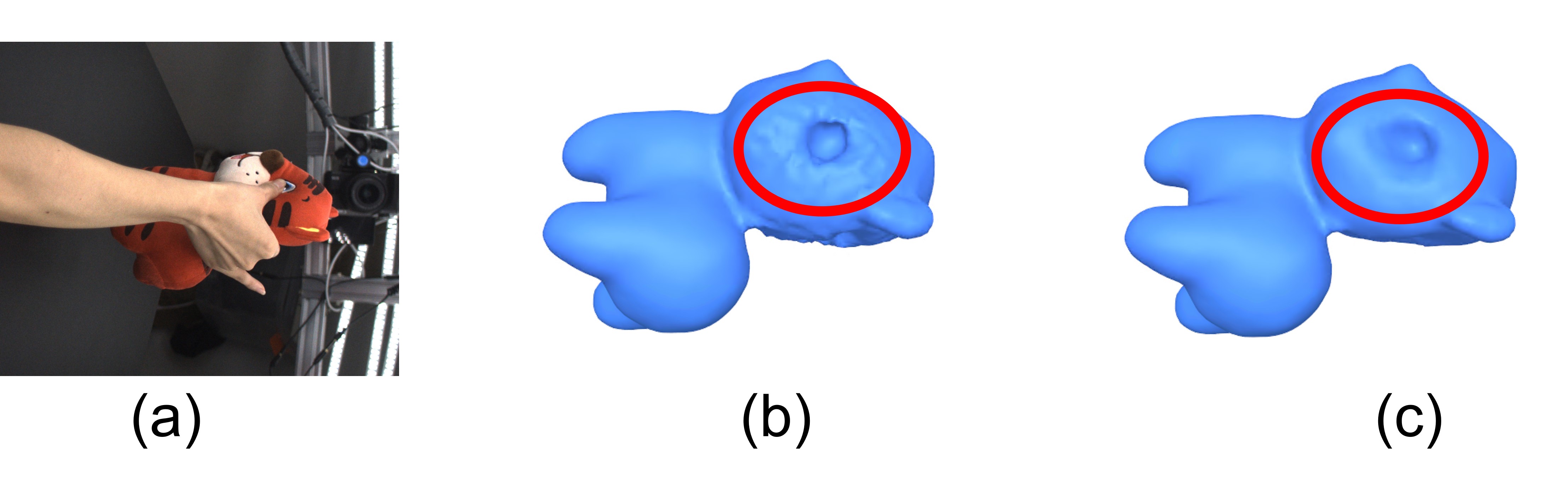}
    \caption{\textbf{Evaluation of the regularization term for object deformation. }(a) Reference frame; (b) Result without the regularization term; (c) Result with the regularization term.}
    \label{fig08_reg_term}
\end{figure}

We perform ablation experiments on the terms in~\equationmk\ref{eqn_object_energy_function}. Regarding the evaluation metrics, We use mIoU to evaluate the quality of deformed object reconstruction results. In addition, \emph{intersection volume} (denoted as Inter. in tables) proposed in~\cite{hasson2019learning} is adopted to evaluate the contact quality between hand and deformed object. The experimental results are shown in~\tablemk\ref{tab04_ablation_object}. Although the term $E_{\text {temp}}$ does not help improve the contact quality, it results in smoother deformation from frame to frame. Satisfactory reconstruction quality can be achieved with all terms introduced. In addition, we show the qualitative evaluation of the regularization term for object deformation in~\figmk\ref{fig08_reg_term}. As shown in ~\figmk\ref{fig08_reg_term} (b), without the regularization term, the transformations of adjacent graph nodes are inconsistent, which leads to unnatural deformation. After introducing the regularization term, a reasonable result is obtained. 

\section{Conclusion}
We construct the first dataset to record the interactive motion of hands and deformable objects to fill the gap in the hand and deformable object datasets.
It captures deformable interactions in multiple interaction forms from 10 perspectives with our multi-view capture system. We propose a method to annotate our captured motion data. The method makes full use of information from various perspectives to reconstruct the accurate hand, and the collaboration between the hand and the object is considered to guide the object pose estimation and the object deformation. 
Through comprehensive evaluation, we demonstrate that our method can reconstruct interactive motions of hands and different deformable objects with high quality. In the future, this dataset can be used for research on hand and deformable object reconstruction.

\noindent\textbf{Limitations and Future Work.} We constructed a dataset containing different forms of deformable interactions, where the main focus is on contact deformation of interacting objects. The interacting objects in our dataset do not have large deformations, such as 180-degree twisting or bending. The proposed hand and deformable object reconstruction method requires the material of deformable objects to be uniform, otherwise our deformation diffusion strategy may not work properly. In the future, we will add large deformations of objects, and consider introducing depth and color information.


{\small
\bibliographystyle{cvm}
\bibliography{cvmbib}
}

\end{document}